\documentclass{article}
\usepackage{algorithm} 
\usepackage[numbers,sort&compress]{natbib}
\usepackage{algpseudocode} 
\usepackage{booktabs}
\usepackage{PRIMEarxiv}
\usepackage[table,xcdraw]{xcolor}
\usepackage[utf8]{inputenc} 
\usepackage[T1]{fontenc}    
\usepackage{hyperref}       
\usepackage{url}            
\usepackage{booktabs}       
\usepackage{amsfonts}       
\usepackage{nicefrac}       
\usepackage{microtype}      
\usepackage{lipsum}
\usepackage{fancyhdr}       
\usepackage{graphicx}       
\graphicspath{{media/}}     
\usepackage{adjustbox}
\usepackage{algorithm} 
\usepackage{algpseudocode}
\usepackage[section]{placeins}
\usepackage{amsmath}
\usepackage{amssymb}
\usepackage{booktabs}
\usepackage{multirow}
\usepackage{multirow,diagbox,tabularx,blindtext}
\usepackage{threeparttable}
\usepackage{color}
\usepackage{float}
\usepackage{tabularray}
\usepackage{longtable}
\usepackage[numbers,sort&compress]{natbib}

\usepackage{lineno}
\title{
Multi-Agent Visual-Language Reasoning for Comprehensive Highway Scene Understanding}

\author{
 Yunxiang Yang, Ningning Xu, Jidong J. Yang$^{*}$ \\
Smart Mobility and Infrastructure Lab\\
College of Engineering, University of Georgia \\
  \texttt{\{yyang117, Ningning.Xu, Jidong.Yang\}@uga.edu} \\
}

\begin{document}
\maketitle

\renewcommand*{\thefootnote}{\fnsymbol{footnote}}
\footnotetext{$^{*}$Corresponding author.}

\begin{abstract}
This paper introduces a multi-agent framework for comprehensive highway scene understanding, designed around a mixture-of-experts strategy. In this framework, a large generic vision-language model (VLM), such as GPT-4o, is contextualized with domain knowledge to generates task-specific chain-of-thought (CoT) prompts. These fine-grained prompts are then used to guide a smaller, efficient VLM (e.g., Qwen2.5-VL-7B) in reasoning over short videos, along with complementary modalities as applicable. The framework simultaneously addresses multiple critical perception tasks, including weather classification, pavement wetness assessment, and traffic congestion detection, achieving robust multi-task reasoning while balancing accuracy and computational efficiency. To support empirical validation, we curated three specialized datasets aligned with these tasks. Notably, the pavement wetness dataset is multimodal, combining video streams with road weather sensor data, highlighting the benefits of multimodal reasoning. Experimental results demonstrate consistently strong performance across diverse traffic and environmental conditions. From a deployment perspective, the framework can be readily integrated with existing traffic camera systems and strategically applied to high-risk rural locations, such as sharp curves, flood-prone lowlands, or icy bridges. By continuously monitoring the targeted sites, the system enhances situational awareness and delivers timely alerts, even in resource-constrained environments.
\end{abstract}


\section{Introduction}

Multimodal foundation models, particularly Vision-Language Models (VLMs), have emerged as powerful artificial intelligence (AI) agents capable of understanding and reasoning across diverse data modalities, such as images, video, audio, and text~\cite{keskar2025evaluating, park2025nuplanqa, luo2024delving}. These models are typically built on transformer-based architectures that integrate information from different modalities into a shared embedding space, enabling the generation of semantically rich, cross-modal representations. A common design pairs a visual encoder (e.g., Vision Transformer) with a language decoder or a unified encoder-decoder architecture, pretrained on large-scale image-text or video-text datasets~\cite{wei2022chain, yang2022prompt}. This large-scale pretraining endows VLMs with strong generalization capabilities across a wide range of vision-and-language tasks, such as image or video captioning, visual question answering (VQA), visual entailment, scene retrieval, among others.

Importantly, the adoption of multimodal foundation models marks a paradigm shift in infrastructure monitoring. Traditional systems often rely on dedicated physical sensors - such as weather stations, embedded pavement sensors, or radar -  which entail significant installation, calibration,  and maintenance costs. In contrast, multimodal models can leverage existing video camera infrastructure (e.g., CCTV cameras) for robust visual reasoning~\cite{gao2025application}. These models are not only capable of assessing environmental attributes (e.g., wet pavement, snow accumulation, visibility reduction) but also recognizing specific hazards such as fallen debris or stalled vehicles. Interactive frameworks like \textit{ SeeUnsafe} exemplify this potential by using VLMs to identify safety-critical events in large-scale traffic video data~\cite{zhang2025when}.

Moreover, recent research has explored multi-task learning paradigms that unify diverse downstream tasks within a shared modeling framework~\cite{luo2024delving}. These foundation models learn transferable representations that span perception, prediction, and decision-making, enabling joint optimization and reducing the need for extensive data annotation. By leveraging shared knowledge across tasks, they offer a holistic and efficient approach to complex transportation scenarios. However, deploying such models in time-critical transportation systems remains challenging due to their large model size and computational overhead. To mitigate these constraints while preserving performance, researchers are exploring strategies such as model distillation, sparse expert routing, and task-specific Chain-of-Thought (CoT) prompting, each aiming to improve inference efficiency and adaptability in resource-constrained environments.

\subsection{Related Work}

While deep learning has enabled considerable progress in traffic scene understanding, most existing approaches remain limited to single-task settings.  This study aims to advance multitask visual understanding for  comprehensive scene interpretation in the transportation domain. Specifically, we focus on jointly modeling and understanding road weather, pavement surface and traffic conditions, enabling a more holistic and robust perception of real-world transportation environments. 

\subsubsection{Road Weather Understanding} 

Accurately assessing weather conditions from roadside or in-vehicle perspectives is critical for maintaining traffic safety and ensuring operational resilience. Conventional approaches are grounded in Numerical Weather Prediction (NWP), which relies on data assimilation of satellite, radar, and in situ observations~\cite{lynch2008origins, zhang2009coupling, bechtold2014representing, geer2017growing, bauer2015quiet}. Recent advancements incorporate deep learning for localized weather understanding—\cite{shi2018weather, Zhen2023CoSupervised} uses Convolutional Neural Networks (CNNs) and Conditional Generative Adversarial Networks (CGANs) for weather classification from road images, while~\cite{qing2018hourly} and~\cite{schmidt2020modeling} utilize LSTM and GAN models for short-term forecasting of solar irradiance and cloud patterns. On the language side, foundation models such as ClimateBERT~\cite{webersinke2021climatebert} and ClimateGPT~\cite{thulke2024climategpt} have been proposed for climate-focused text understanding.

However, these approaches either focus exclusively on visual inputs or treat weather as a standalone forecasting problem. They often lack real-time road-level granularity or integration with traffic scene contexts, limiting their utility for real-time hazard detection and warning. In contrast, VLMs offer the ability to infer weather directly from traffic videos and reason about its impact on safety conditions (e.g., reduced visibility, road surface conditions), thus closing the gap between meteorological modeling and transportation decision-making.

\subsubsection{Pavement Wetness Assessment}

Timely detection of pavement wetness is essential for highway safety and operations, as surface water significantly reduces tire–pavement friction and increases the likelihood of hydroplaning, particularly at high speeds. These conditions not only elevate crash risk but also complicate traffic management and roadway maintenance decisions. Traditional approaches rely on embedded sensors or road weather stations, which are accurate but sparse and expensive to maintain~\cite{khan2022weather}. Modern deep learning systems have applied CNNs and segmentation networks to RGB or infrared images to classify wet, dry, snowy, or icy surfaces~\cite{chandra2022deep, yuan2021road}. Acoustic sensing systems, such as those by~\cite{kalliris2019machine}, utilize tire-road interaction sounds to estimate surface wetness using Support Vector Machines and logistic regression models.

Recent webcam-based methods~\cite{khan2022weather} leverage pre-trained CNNs (e.g., ResNet18) to identify pavement conditions from roadside imagery. Thermal imaging has also been explored to detect sub-surface anomalies and transient wetness features~\cite{chandra2022deep}. Hybrid approaches like RMSDC~\cite{elwahsh2023deep} fuse temporal sensor data using ConvLSTM for robust, interpretable predictions. Despite the progress, these methods are typically static, infrastructure-specific, and lack adaptability across domains. They often require extensive re-labeling or fine-tuning when deployed in new regions or under different weather conditions.

\subsubsection{Congestion Analysis}

Traffic congestion detection is another domain where deep learning methods has gradually replaced traditional models. Hybrid CNN-LSTM architectures~\cite{mihaita2020traffic} and encoder-based deep networks~\cite{liu2023highway} have been developed to model spatio-temporal traffic dynamics from loop detectors and speed sensors. Vision-based methods have also been applied, enabling real-time congestion classification directly from traffic video feeds~\cite{chakraborty2018traffic}.

However, most of these models operate as task-specific detectors trained on specific datasets. They lack semantic understanding and struggle with context-sensitive reasoning (e.g., distinguishing construction-induced slowdown from other congestion scenarios). Recent reviews~\cite{azfar2024deep} advocate for more explainable and generalizable frameworks. Vision-language models have  shown promise in this direction, offering semantic alignment between scene content and user-defined queries, enabling interpretable diagnostics and causality analysis of congestion~\cite{keskar2025evaluating, zhang2025when}.


Despite rapid advancements, most existing approaches remain task-specific, requiring frequent retraining and exhibiting limited adaptability to new scenarios. To overcome these limitations, we propose a VLM-driven framework for comprehensive highway scene understanding. Our approach employs a mixture-of-agents strategy, in which a large generic VLM (e.g., GPT-4o) generates fine-grained CoT prompts tailored to three core perception tasks spanning the environment (weather classification), roadway (pavement wetness assessment), and traffic (congestion detection). These prompts are then used to guide a smaller, computationally efficient VLM (e.g., Qwen2.5-VL-7B) in reasoning over short video inputs, enabling robust and scalable multi-task scene understanding. 

To support rigorous evaluation, we curated three dedicated datasets, each aligned with one of the target tasks. Notably, the pavement wetness dataset is multi-modal, combining video footage with road weather station data to demonstrate the advantages of multimodal reasoning. Experimental results validate the effectiveness of our approach and highlight the potential of collaborative VLM agents for understanding comprehensive highway scenes. 

\section{Dataset}
We collected the publicly accessible traffic video data from the states of Georgia, Virginia, and California. Camera locations were strategically chosen to cover urban, suburban, mountain, and coastal regions, ensuring a diverse set of highway scenes.

\subsection{Category Definition}
For weather classification, we focus on three primary conditions: clear (including sunny and cloudy, with no precipitation), rainy, and snowy. For pavement wetness level assessment, we defined seven categories aligned with corresponding weather conditions: dry, rainy fully wet, rainy partially wet, rainy flooded, snowy fully wet, snowy partially wet, and snowy wet with icy warning. A detailed description of each category is provided in Table \ref{tab:pavement_wetness}. 

\begin{table}[!ht]
    \caption{Pavement wetness level definitions and visual cues}
    \label{tab:pavement_wetness}
    \begin{center}
        \begin{tabular}{l p{11cm}}
            \hline
            \textbf{Category} & \textbf{Key Visual and Contextual Cues} \\
            \hline
            Rainy fully wet & Uniformly dark and glossy surface with consistent reflections. Tire sprays are visible across lanes; vehicles often leave moderate water trails. \\
            Rainy partially wet & Mixed appearance with wet patches and dry zones. Water sprays are intermittent or limited to certain lanes. Some vehicles show water trails, others do not. \\
            Rainy flooded & Standing or pooling water is clearly visible. Vehicles generate large water splashes and long, wide spray plumes. Water trails are thick and persistent. \\
            Dry & Light-colored, matte surface with no visible moisture or reflections. No tire  sprays or water trails; vehicle movement is clean and uninhibited. \\
            Snowy fully wet & Entire surface is dark and wet from melted snow. Slush may appear near curbs or median dividers. Tire water spray may be visible. No snow patches. \\
            Snowy partially wet & Uneven surface with a mix of wet, dry, or slushy zones. Residual snow or damp spots are visible. Minimal and inconsistent water sprays. \\
            Snowy wet with icy warning & Surface has a faint shine or frosty gloss suggesting potential black ice. This is often coupled with low temperature and high humidity conditions. Sprays are minimal or absent. Vehicles may move slowly with extra caution. \\
            \hline
        \end{tabular}
    \end{center}
\end{table}

For congestion detection, we grouped traffic flow conditions into two categories: congested and unobstructed. The detailed descriptions are given in Table \ref{tab:traffic_flow_definitions}. The distribution of video clips across weather conditions is presented in Table \ref{tab:weather_summary}.

\begin{table}[!ht]
    \caption{Traffic flow condition definitions and visual cues}
    \label{tab:traffic_flow_definitions}
    \begin{center}
        \begin{tabular}{l p{11cm}}
            \hline
            \textbf{Category} & \textbf{Key Visual and Contextual Cues} \\
            \hline
            Congested & Lanes are visibly full or nearly full of vehicles with minimal open space. Vehicle spacing is tight, making lane changes difficult. Motion is uneven; cars exhibit stop–go patterns, frequent surging and braking, or shock-wave movements. Multiple vehicles display delayed following, indicating disrupted flow. \\
            Unobstructed & Traffic flows smoothly at or near posted speeds. Vehicles are evenly spaced. Motion is steady with little to no deceleration. The road appears open with no apparent disruptions to traffic flow. \\
            \hline
        \end{tabular}
    \end{center}
\end{table}


\begin{table}[!ht]
    \caption{Summary of the weather video dataset}
    \label{tab:weather_summary}
    \begin{center}
        \begin{tabular}{l c}
            \hline
            \textbf{Weather Condition} & \textbf{Total Videos} \\
            \hline
            Clear  & 66 \\
            Snowy  & 21 \\
            Rainy  & 73 \\
            \hline
            \textbf{Total} & \textbf{160} \\
            \hline
        \end{tabular}
    \end{center}
\end{table}

\begin{figure}[!ht]
    \centering
    \includegraphics[width=0.8\linewidth]{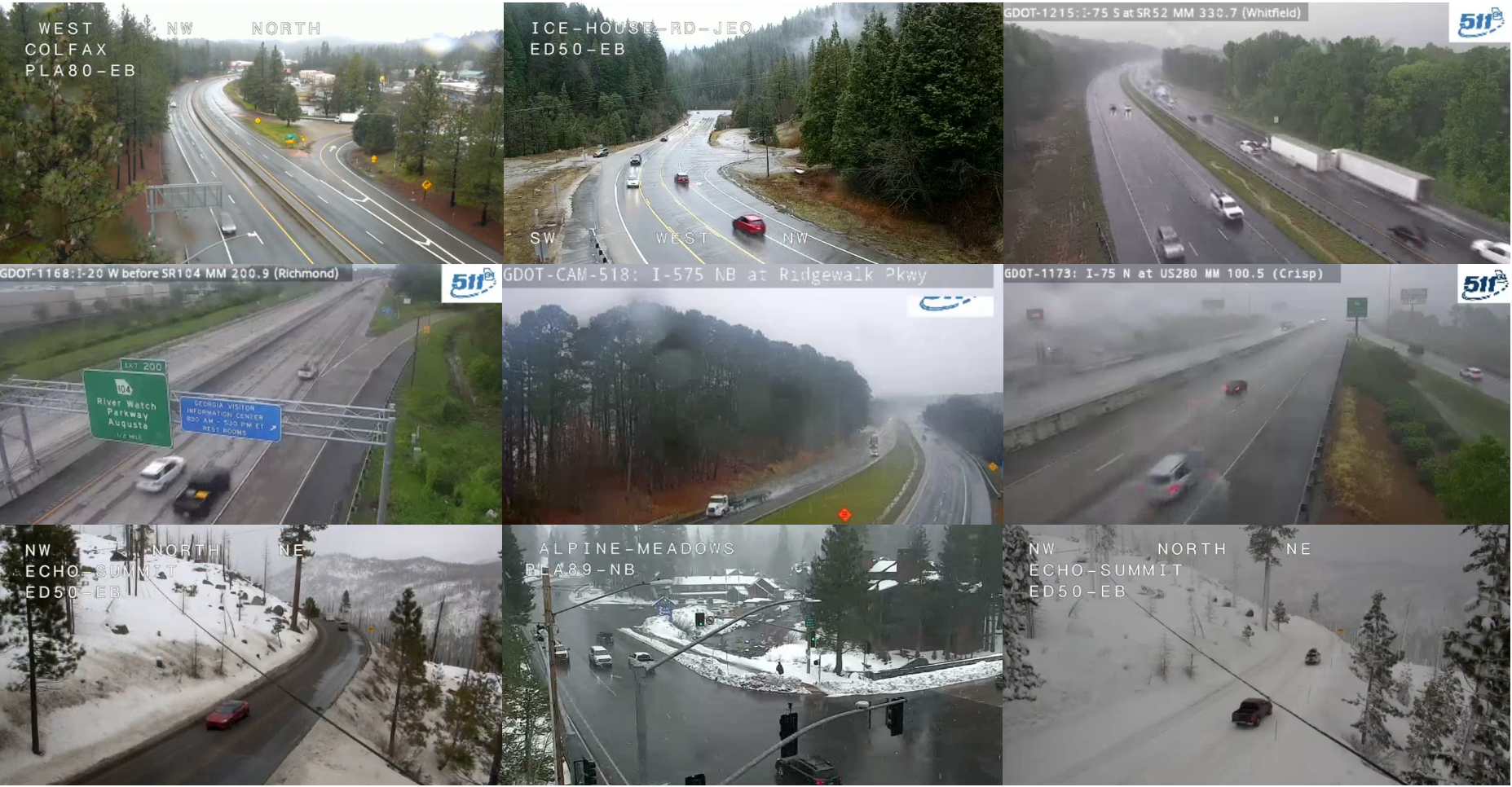}
    \caption{Examples from the pavement wetness assessment dataset. The leftmost column: 1st and 2nd rows – rainy partially wet; 3rd row – snowy partially wet. The middle column: 1st and 2nd rows – rainy fully wet; 3rd row – snowy fully wet. The rightmost column: 1st and 2nd rows – rainy flooded; 3rd row – snowy wet with icy warning.}
    \label{fig:wetness_example}
\end{figure}

In addition to the video data, we collected information from the nearest road weather stations, resulting in a multimodal dataset. Depending on the availability of sensor readings at the time of data collection, we distinguish between partial and full multimodal data. Partial multi-modal data includes: date/time, current weather, weather precipitation, temperature high/low, elevation. In contrast, full multimodal data provides more detailed environmental context, including: date/time, relative humidity, wind speed/direction, air/surface temperature, visibility, dew point temperature, surface condition, and precipitation. For our downstream tasks, we primarily leverage this multimodal data for pavement wetness assessment under snowy conditions, where cross-modal reasoning provides the greatest benefit. 
A detailed summary of this dataset is shown in Table \ref{tab:dataset_summary}, and some examples are given in Figure \ref{fig:wetness_example}.

\begin{table}[!ht]
    \caption{Summary of the pavement wetness video dataset}
    \label{tab:dataset_summary}
    \begin{center}
        \begin{tabular}{l c c}
            \hline
            \textbf{Category} & \textbf{Total Videos} & \textbf{Multi-Modal Data Type} \\
            \hline
            Rainy partially wet         & 51  & 46 Full, 5 Partial \\
            Rainy fully wet             & 73  & 68 Full, 5 Partial \\
            Rainy flooded               & 18  & Full only \\
            Snow partially wet          & 5   & Partial only \\
            Snow fully wet              & 9   & Partial only \\
            Snow wet with icy warning   & 21  & Partial only \\
            Sunny dry                   & 66  & Full only \\
            \hline
            \textbf{Total}              & \textbf{243} & \\
            \hline
        \end{tabular}
    \end{center}
\end{table}

\begin{figure}[!ht]
    \centering
    \includegraphics[width=0.8\linewidth]{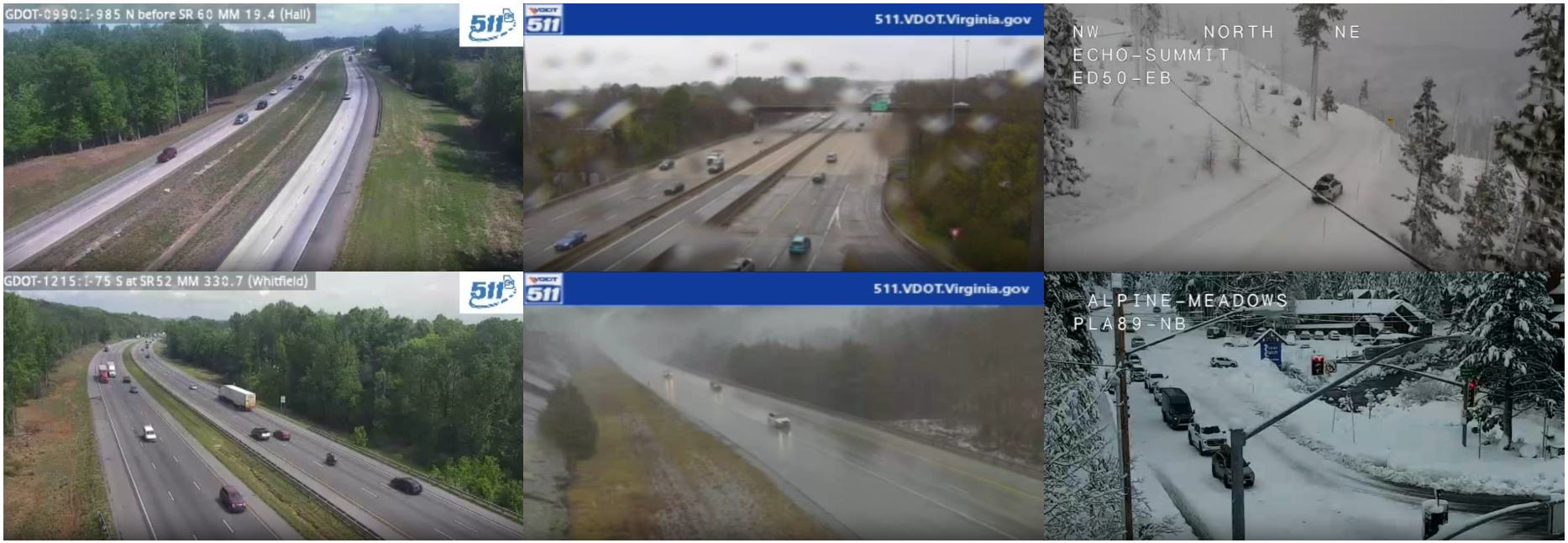}
    \caption{Examples from the road weather classification dataset. The leftmost column: clear weather conditions. The middle column: rainy conditions. The rightmost column: snowy conditions.}
    \label{fig:weather_example}
\end{figure}

It is important to note that for congestion analysis, rather than assigning a single label (e.g., congested or unobstructed) to an entire road segment within a video, we explicitly specify the traffic direction of the segment. This distinction accounts for the possibility of direction-dependent traffic patterns. Specifically, inbound refers to vehicles moving toward the traffic camera, while outbound refers to those moving away from it. 

\begin{table}[!ht]
    \caption{Summary of the traffic congestion video dataset}
    \label{tab:congestion_summary}
    \begin{center}
        \begin{tabular}{l l c}
            \hline
            \textbf{Congestion Level} & \textbf{Direction} & \textbf{Total Videos} \\
            \hline
            {Congested} 
            & Inbound  & 10 \\
            & Outbound & 18 \\
            \hline
            {Unobstructed} 
            & Inbound  & 20 \\
            & Outbound & 16 \\
            \hline
            \textbf{Total} &  & \textbf{64} \\
            \hline
        \end{tabular}
    \end{center}
\end{table}

\begin{figure}[!ht]
    \centering
    \includegraphics[width=0.7\linewidth]{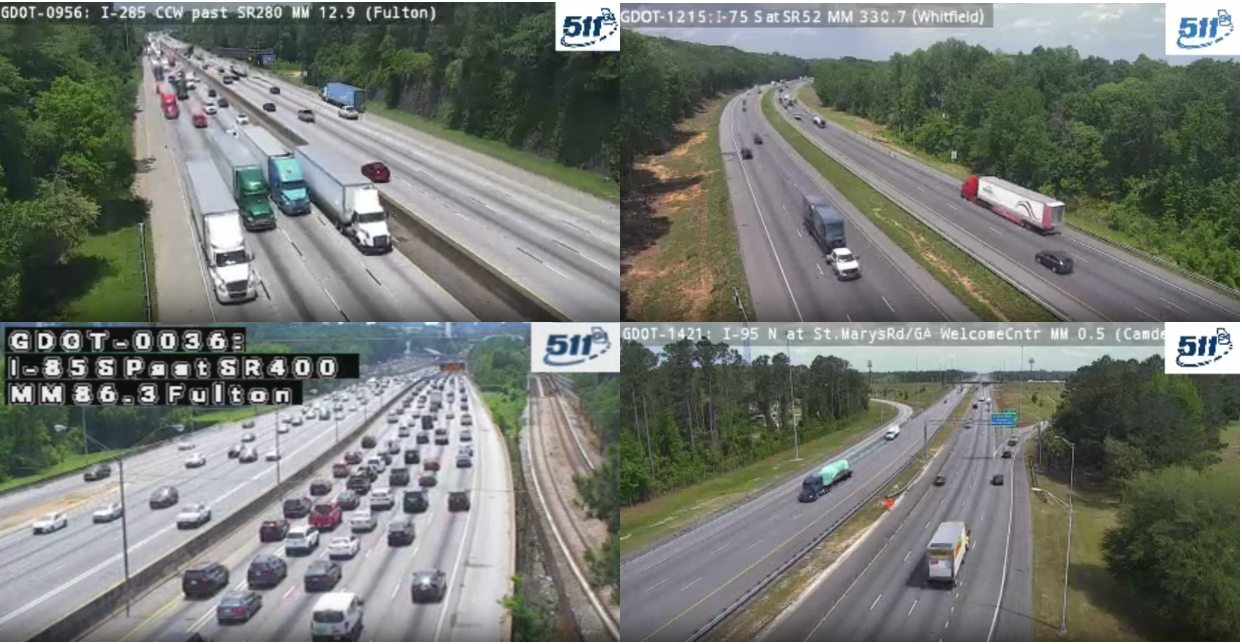}
    \caption{Examples from the congestion analysis dataset. The left column: 1st row – inbound congested; 2nd row – outbound congested. The right column: 1st row – inbound unobstructed; 2nd row – outbound unobstructed.}
    \label{fig:congestion_example}
\end{figure}

\section{Method}

\begin{figure}[!ht]
    \centering
    \includegraphics[width=1\linewidth]{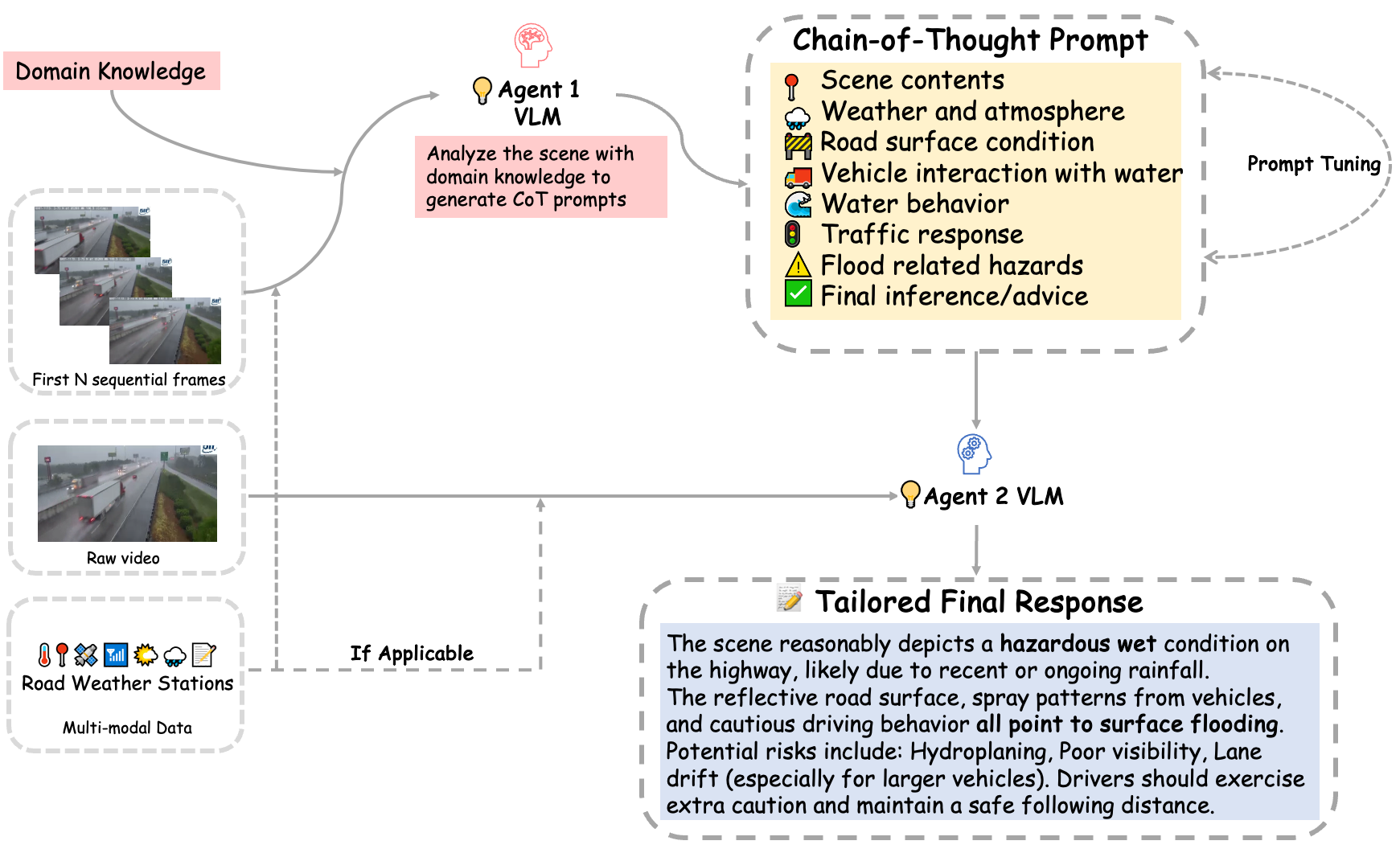}
    \caption{The multi-agent framework for highway scene understanding.}
    \label{fig:framework}
\end{figure}

This section introduces our proposed method, which leverages multiple agents to understand traffic scenes (weather, pavement wetness, congestion, etc.). The process begins by extracting sequential frames from a video input to retain temporal dynamics. An initial prompt, incorporating relevant domain knowledge, is constructed and provided to a VLM, referred to as Agent 1. In our experiment, we use GPT-4o \cite{openai2023gpt4} in this role. Agent 1 analyzes a scene based on the initial prompt and generates a detailed Chain-of-Thought (CoT) \cite{wei2022chain} prompt that systematically address multiple aspects of the scene from the surrounding environment to vehicles. Depending on the downstream task, such as pavement wetness assessment under snowy conditions, multimodal data can also be ingested to enhance reasoning. Prompt tuning \cite{yang2022prompt} is also applied to ensure accurate description of the scene is aligned with human observations and domain knowledge. The refined CoT prompt is then passed to Agent 2, which perform inference directly on video inputs and, if applicable, with associated multi-modal data. For Agent 2, we use QWEN 2.5-VL-7B \cite{bai2025qwen25vl}, which performs CoT-guided reasoning to generate the final output. This multi-agent framework is illustrated in Figure \ref{fig:framework}.

\subsection{CoT Prompts Design and Generation}
\begin{figure}[!ht]
    \centering
    \includegraphics[width=1\linewidth]{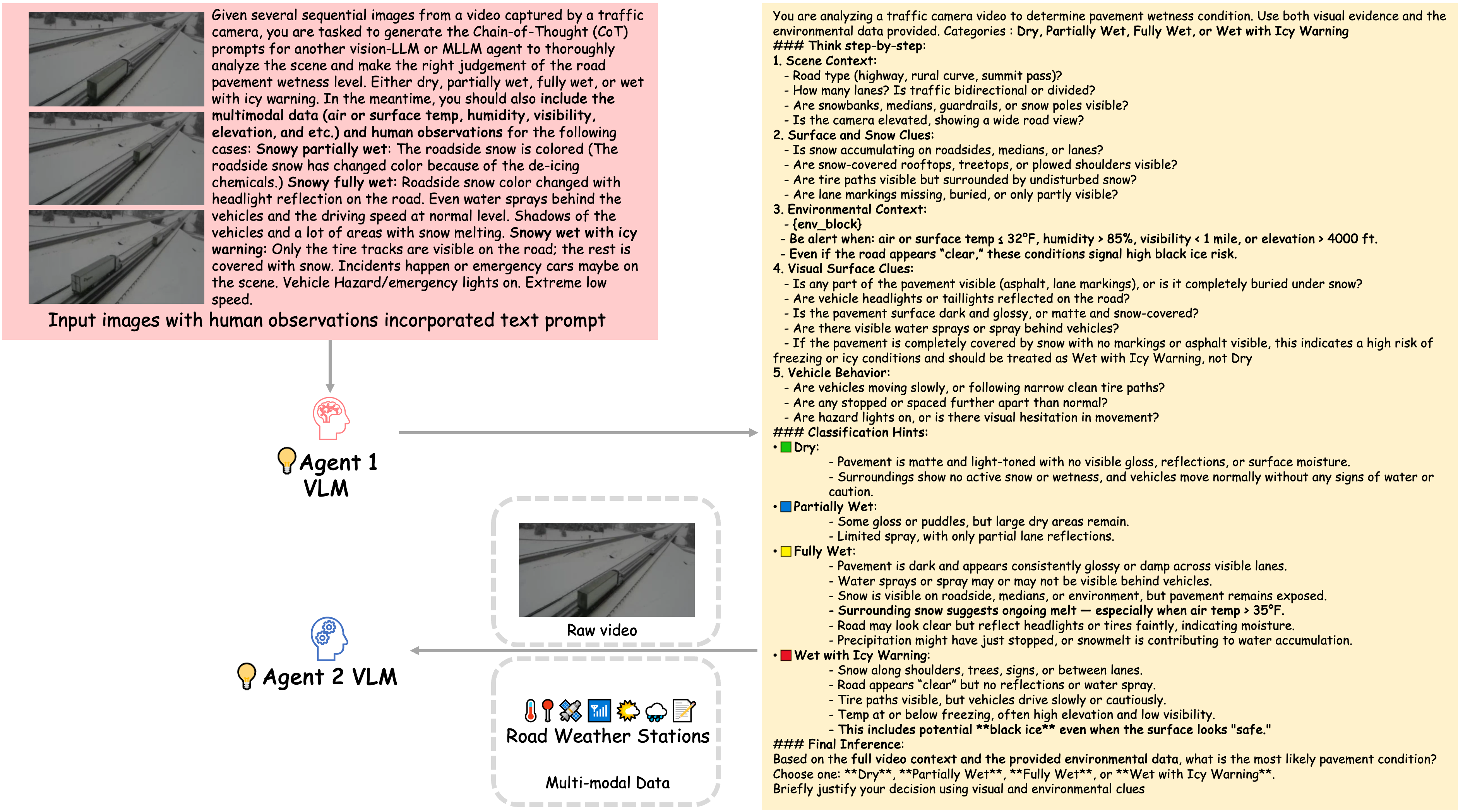}
    \caption{Example of CoT prompt generation for pavement wetness level assessment under snowy conditions.}
    \label{fig:snowy_cot}
\end{figure}

As introduced previously, the definitions used as part of the prompt for Agent 1 (e.g., GPT-4o) integrate human observations and transportation domain knowledge tailored to specific types of scenes. For example, in the case of pavement wetness assessment under snowy conditions, these definitions are combined with sequential video frames and multimodal data (e.g., temperature high/low, humidity, wind speed/direction, dew-point temperature, etc.), and passed to Agent 1 to analyze the scene and acquire the fine-grained CoT prompt. This process is illustrated in Figure \ref{fig:snowy_cot}.

\begin{figure}[!ht]
    \centering
    \includegraphics[width=1\linewidth]{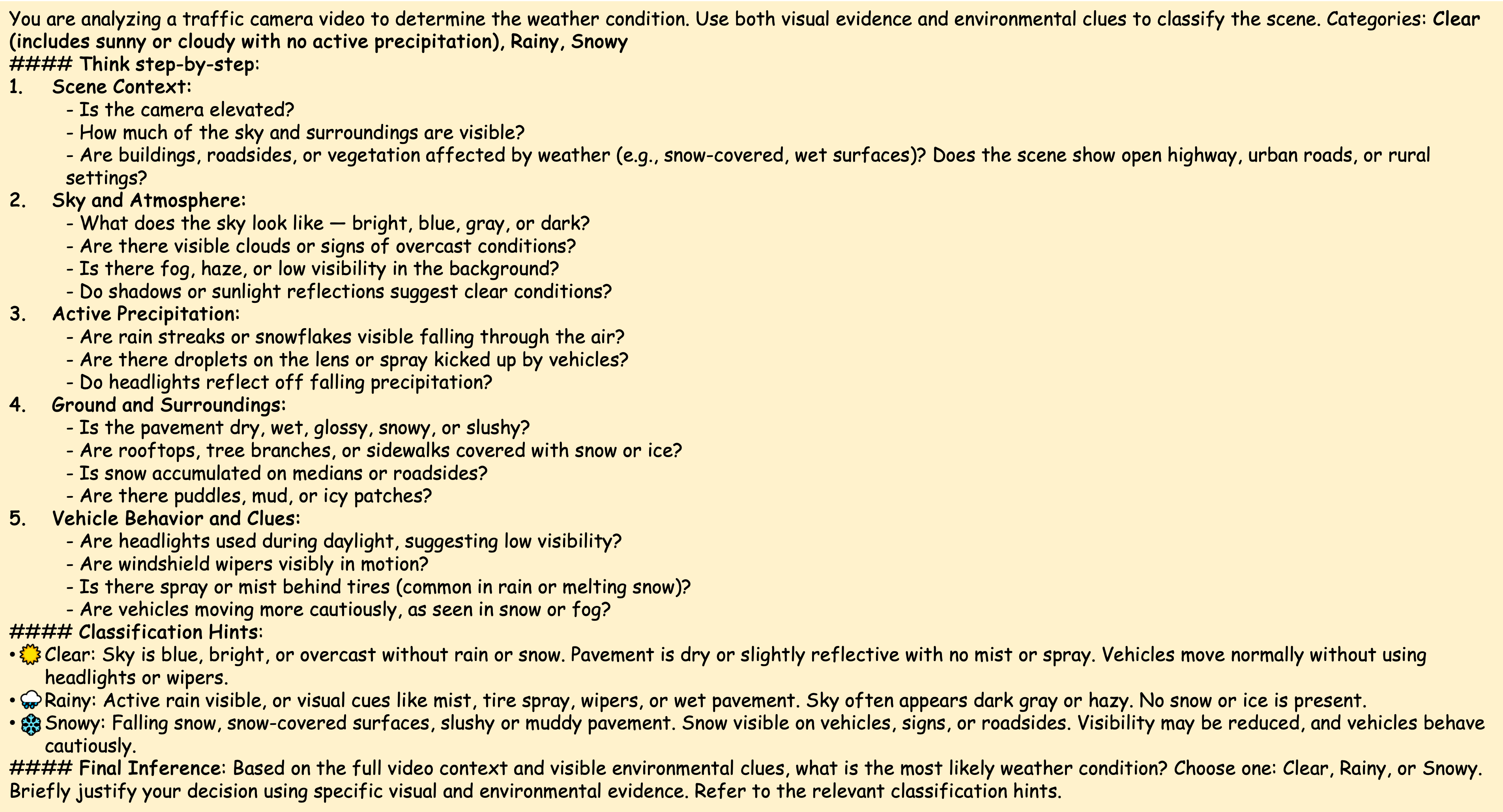}
    \caption{Generated CoT prompt for road weather understanding.}
    \label{fig:cot_weather}
\end{figure}

\begin{figure}[!ht]
    \centering
    \includegraphics[width=1\linewidth]{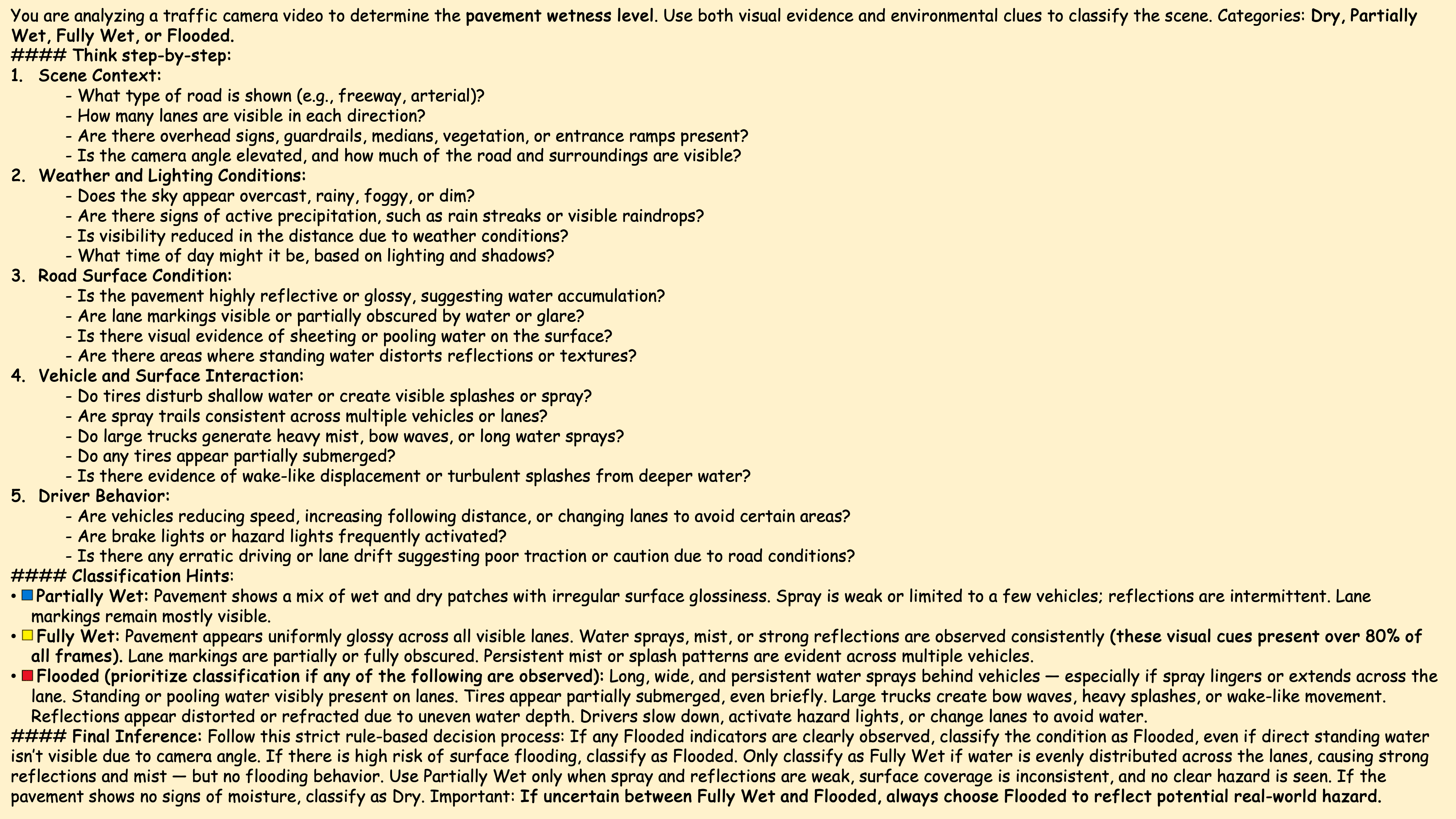}
    \caption{Generated CoT prompt for pavement wetness level assessment under rainy conditions.}
    \label{fig:cot_wetness}
\end{figure}

\begin{figure}[htbp!]
    \centering
    \includegraphics[width=1\linewidth]{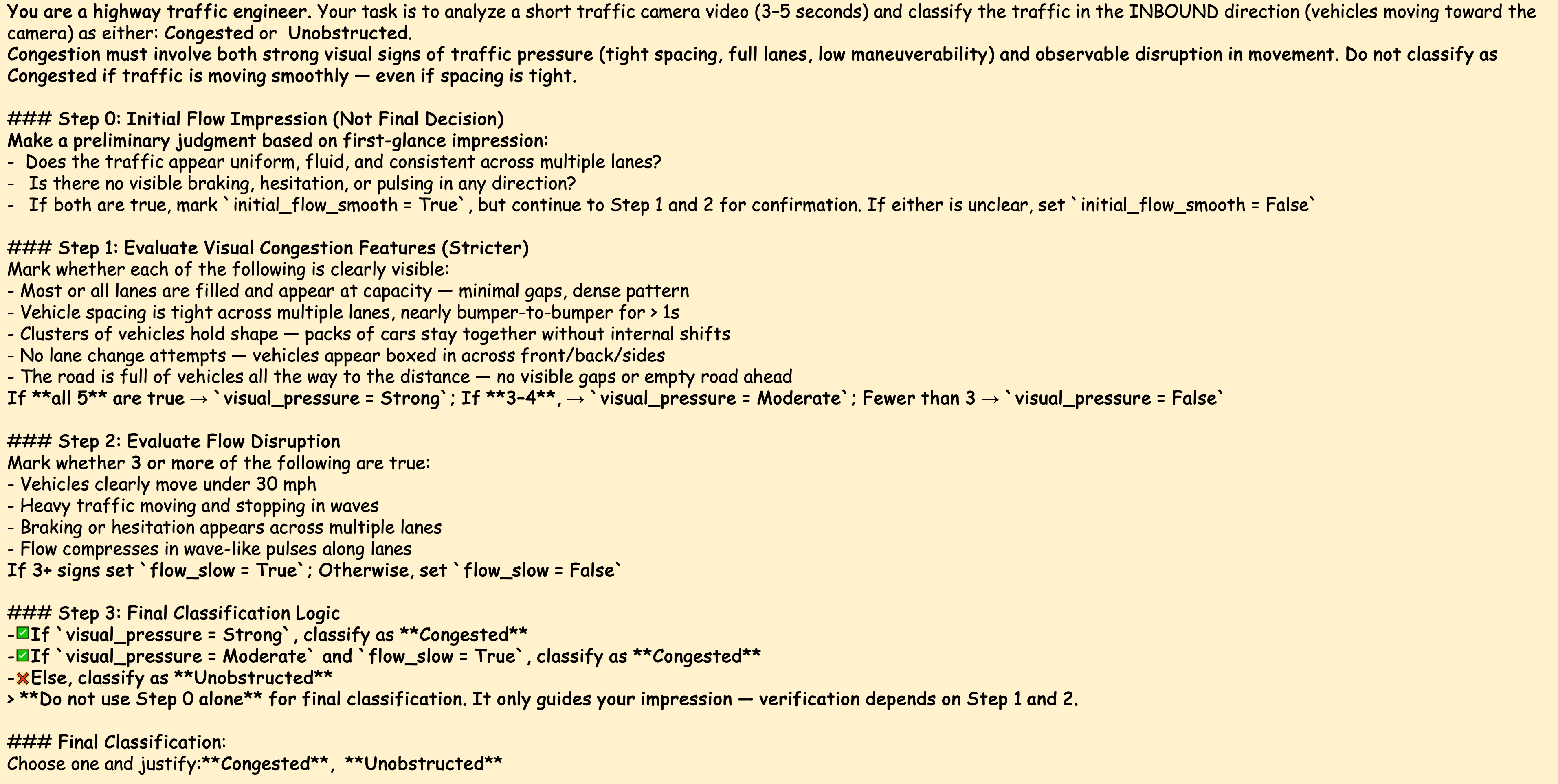}
    \caption{Generated CoT prompt for congestion analysis.}
    \label{fig:cot_congestion}
\end{figure}

Following a similar process, we generate CoT prompts for different downstream tasks. Specifically, Figures \ref{fig:cot_weather}, \ref{fig:cot_wetness}, and \ref{fig:cot_congestion} show the generated CoT prompts by Agent 1 (GPT-4o) for road weather understanding, pavement wetness level assessment, and congestion analysis, respectively. To improve classification accuracy for pavement wetness levels, We introduce a threshold for identifying "fully wet" surface condition, which is defined as  over 80\% of vehicles per frame consistently have water sprays, mist or strong reflections, which helps the model better distinguish partially wet and fully wet. 

We prioritize the "flooded" condition whenever clear visual cues are present (see Figure \ref{fig:cot_wetness}) and explicitly instruct the model: "If uncertain between fully wet and flooded, always choose flooded to reflect the potential real-world hazard." This safety-oriented directive biases the model toward conservative decision-making, helping ensure reliable detection of flooded scenes.

For congestion analysis, we design tailored strategies to address the following challenges: (1) For inference efficiency, all video clips have a short length of 4-7 seconds; the acute signs of traffic congestion may not be obvious within a short video clip. (2) Each traffic camera has a different height and angle, which brings different visual perspectives on traffic flow. (3) During normal peak hours (i.e., without accidents or road closures), the space headway can be short, but vehicles can still move relatively fast. (4) VLMs show limited ability to accurately assess traffic flow speed or other related dynamic features, especially in such short video clips. Our proposed solution introduces a two-variable gating logic that incorporates  \textit{visual\_pressure} and \textit{flow\_slow}, with the initial flow impression serving as a soft flag to provide contextual bias (e.g., "The flow appears smooth, but let me verify"). This design reduces over-reliance on visual cues alone for congestion detection. We further define three levels of \textit{visual\_pressure} (i.e., strong, moderate, and weak) based on the number of visual congestion features identified. In parallel, the \textit{flow\_slow} variable is evaluated as a Boolean flag (‘true’ or ‘false’) depending on evidence of flow disruption. The CoT prompt implementing this gating logic is illustrated in Figure \ref{fig:cot_congestion}.

\section{Experiments and results}
\begin{figure}[!ht]
    \centering
    \includegraphics[width=0.8\linewidth]{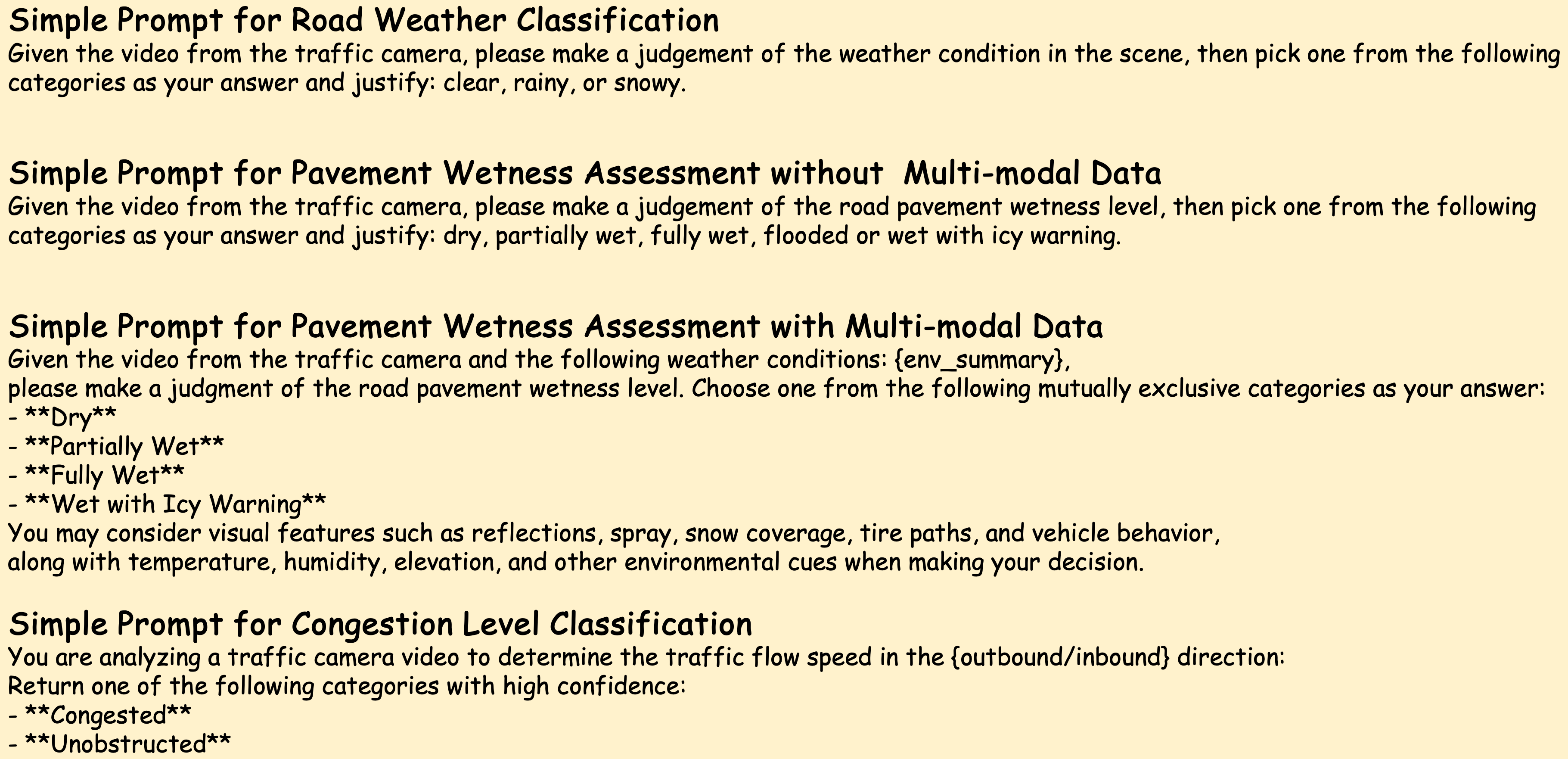}
    \caption{Simple prompts.}
    \label{fig:simple_prompt}
\end{figure}

\begin{figure}[!ht]
    \centering
    \includegraphics[width=1\linewidth]{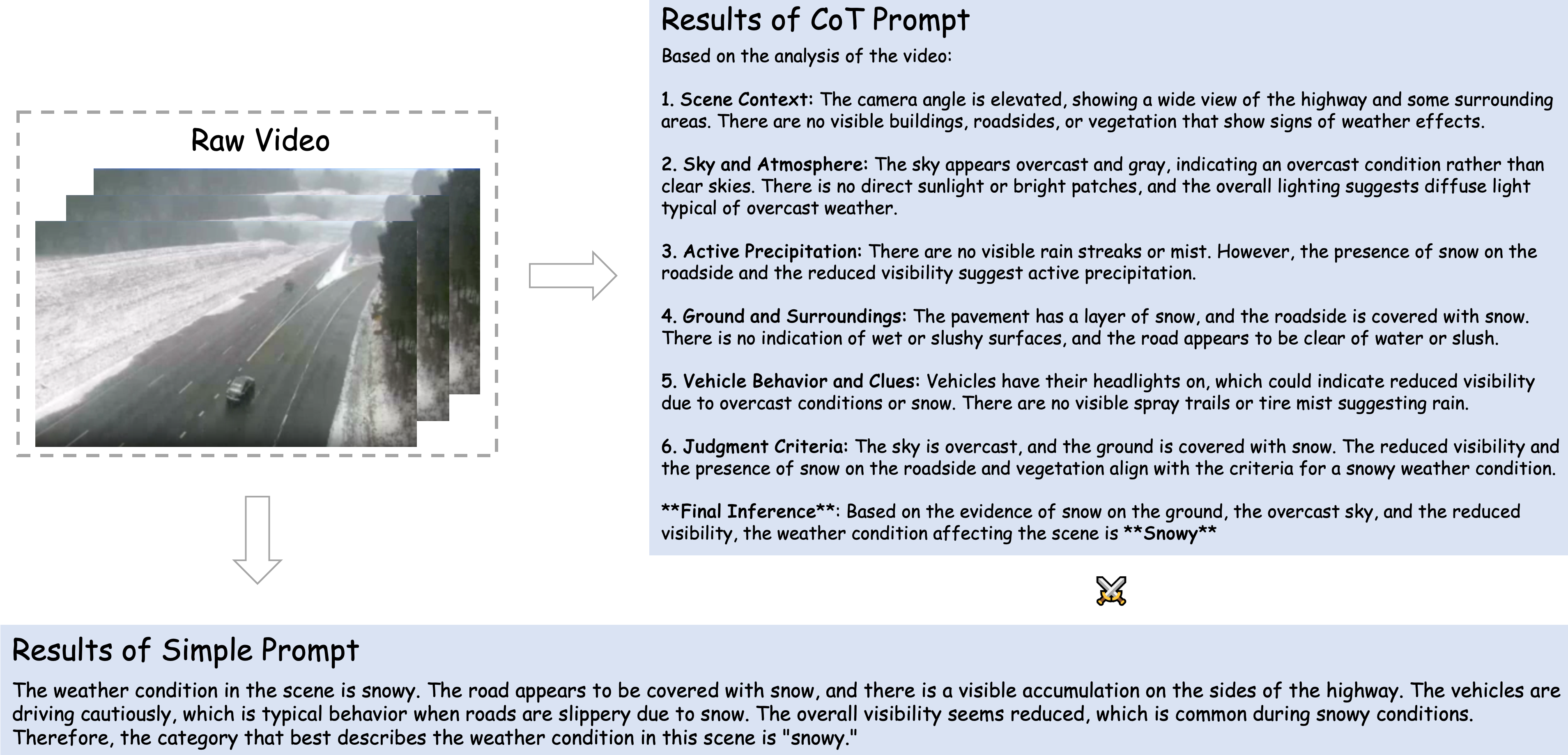}
    \caption{Results of road weather classification via CoT prompt.}
    \label{fig:weather_result}
\end{figure}

We conducted extensive experiments to compare the performance of simple prompts (see Figure \ref{fig:simple_prompt}) versus CoT prompts presented in the preceding section. Exemplar results are shown in Figures \ref{fig:weather_result} - \ref{fig:wetness_snowy_icy_simple_mm_comparison}.  

\subsection{Road Weather Classification}
Road weather classification is relatively straightforward using either simple or CoT prompts, as VLMs can effectively leverage the abundant visual cues present in most conditions. For clear and snowy weather, performance differed only slightly: the model achieved over 90\% accuracy for clear conditions and perfect accuracy for snowy conditions, where the visual evidence of snow is highly distinctive. In contrast, performance declined under rainy conditions, with some rainy scenes misclassified as clear when the visual cues were insufficiently pronounced. 

\begin{table}[!ht]
    \caption{Accuracy comparison between simple and CoT prompts for weather understanding using QWEN 2.5-VL 7B model}
    \label{tab:ablation_weather}
    \begin{center}
        \begin{tabular}{l c c}
            \hline
            \textbf{Weather Condition} & \textbf{Simple Prompt (\%)} & \textbf{CoT Prompt (\%)} \\
            \hline
            Clear  & 92.42  & 95.45 (\textbf{↑3.03}) \\
            Snowy  & 100.00 & 100.00 (\textbf{—}) \\
            Rainy  & 73.24  & 71.23 (\textbf{↓2.01}) \\
            \hline
        \end{tabular}
    \end{center}
\end{table}

\subsection{Pavement Wetness Level Classification}
\begin{figure}[!ht]
    \centering
    \includegraphics[width=1\linewidth]{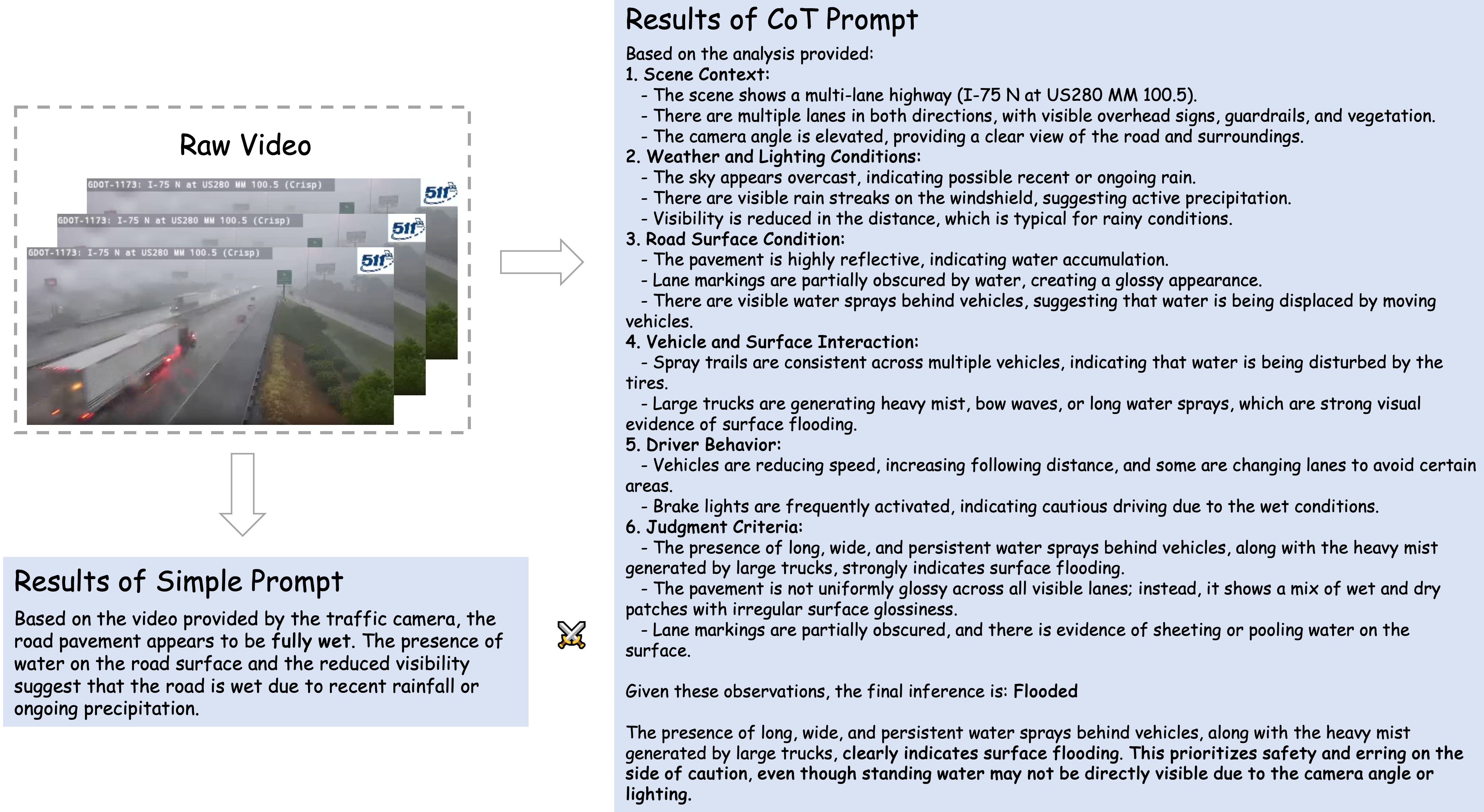}
    \caption{Results of pavement wetness level classification via CoT prompt and simple prompt under flooded condition.}
    \label{fig:wetness_flooded_result}
\end{figure}

First of all, prompting the model via CoT or simple prompt can result in significantly different results and accuracies. A simple case in Figure \ref{fig:wetness_flooded_result} shows the prediction results under a flooded condition. The simple prompt appears to make a judgment based on the first glance without careful attention. In contrast, The CoT prompt instructs the model to look through several aspects, from the scene to road surface, weather, vehicle and driver behaviors, particularly inspect the existence of standing water. Eventually the model tags the video as flooded with the evidence of persistent tire splash-and-spray effects behind vehicles. 

\begin{figure}[!ht]
    \centering
    \includegraphics[width=1\linewidth]{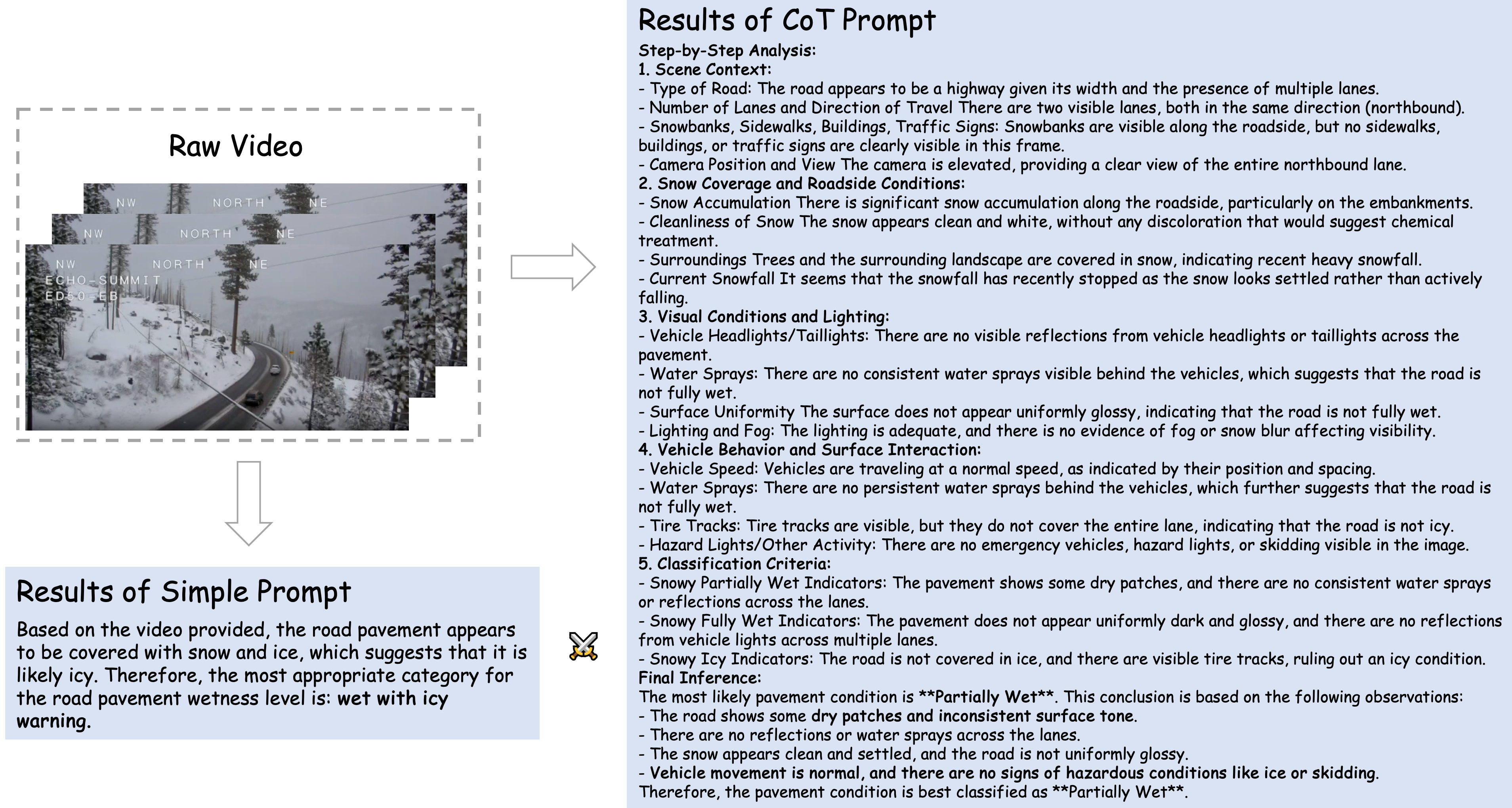}
    \caption{Results of pavement wetness level classification via CoT prompt and simple prompt under snowy partially wet condition.}
    \label{fig:wetness_snowy_partially_wet_result}
\end{figure}

Snowy partially wet condition presents a particularly challenging and ambiguous case since it can easily be misinterpreted as fully wet or icy, even by human observers without careful inspection. This is where the CoT approach outperforms a simple prompt. As shown in Figure \ref{fig:wetness_snowy_partially_wet_result}, the model, guided by a well-crafted CoT prompt, arrived at the correct assessment by systematically examining multiple wetness indicators and ruling out conditions that did not meet the criteria. 

\subsection{Congestion Level Classification}
\begin{figure}[!ht]
    \centering
    \includegraphics[width=1\linewidth]{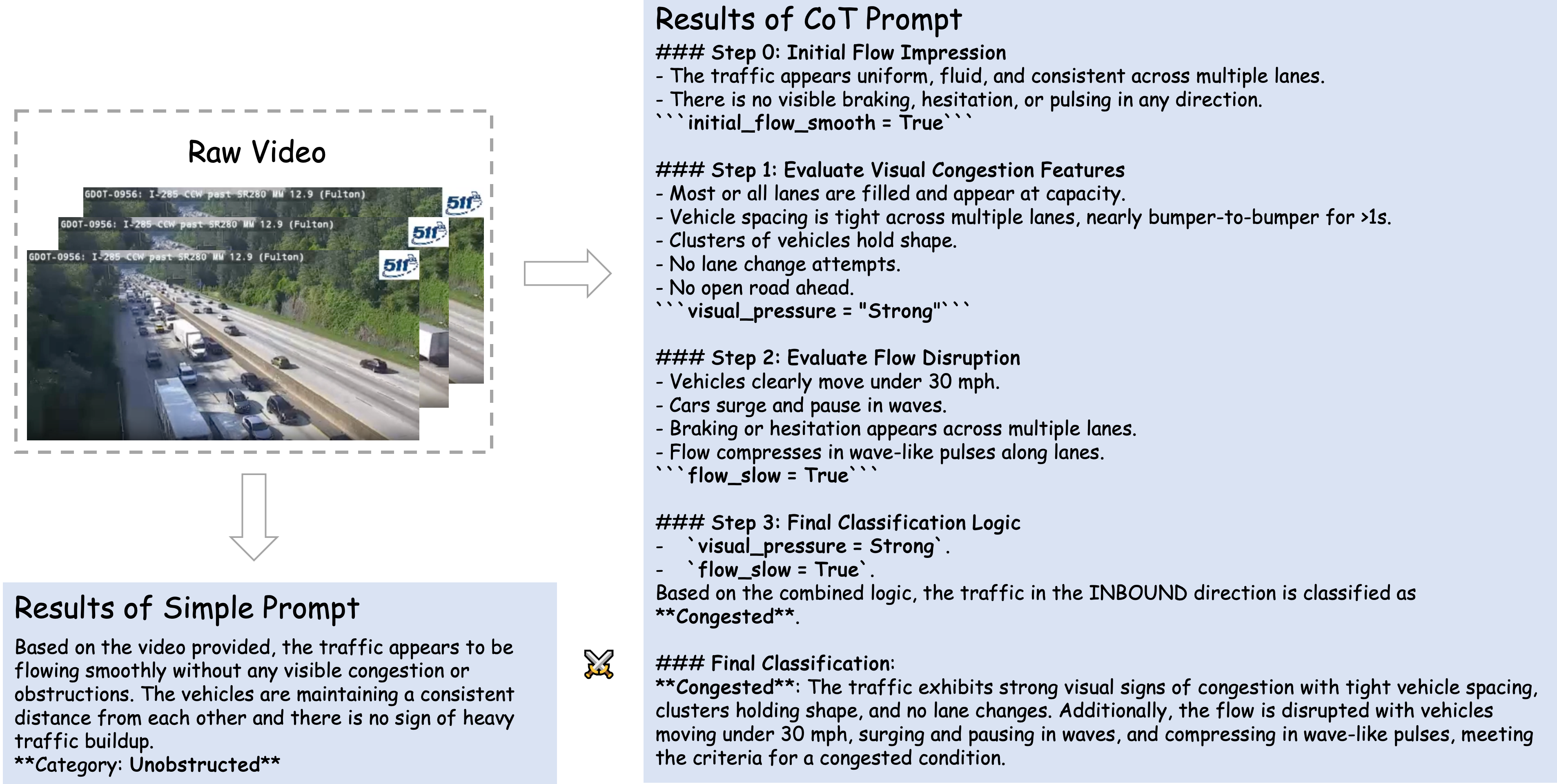}
    \caption{Results of congestion level classification via CoT prompt and simple prompt under congested condition.}
    \label{fig:congestion_result}
\end{figure}

Figure \ref{fig:congestion_result} illustrates that even when the initial flow appears smooth, the model can still identify the congestion by leveraging the values of \textit{visual\_pressure} and \textit{flow\_slow}. This two-variable gating logic, combined with an initial soft flag, significantly enhances accuracy by striking a balance between the VLM model's strengths, such as identifying visual congestion features, and its limitations, particularly in estimating traffic flow speed.  In contrast, the simple prompt relied on the initial impression of traffic flow without further verification, leading to an incorrect judgment.

Table \ref{tab:ablation_congestion} presents improved performance for different directions and cases after implementing our carefully designed CoT prompt.

\begin{table}[!ht]
    \caption{Accuracy comparison between simple and CoT prompts for congestion analysis using QWEN 2.5-VL 7B model}
    \label{tab:ablation_congestion}
    \begin{center}
        \begin{tabular}{l l c c}
            \hline
            \textbf{Congestion Level} & \textbf{Direction} & \textbf{Simple Prompt (\%)} & \textbf{CoT Prompt (\%)} \\
            \hline
            {Congested} 
            & Inbound  & 40.00  & 100.00 (\textbf{↑60.00}) \\
            & Outbound & 72.22  & 100.00 (\textbf{↑27.78}) \\
            \hline
            {Unobstructed} 
            & Inbound  & 60.00  & 80.00 (\textbf{↑20.00}) \\
            & Outbound & 81.25  & 100.00 (\textbf{↑18.75}) \\
            \hline
        \end{tabular}
    \end{center}
\end{table}

\subsection{Multimodal Reasoning}
\begin{figure}[!ht]
    \centering
    \includegraphics[width=0.95\linewidth]{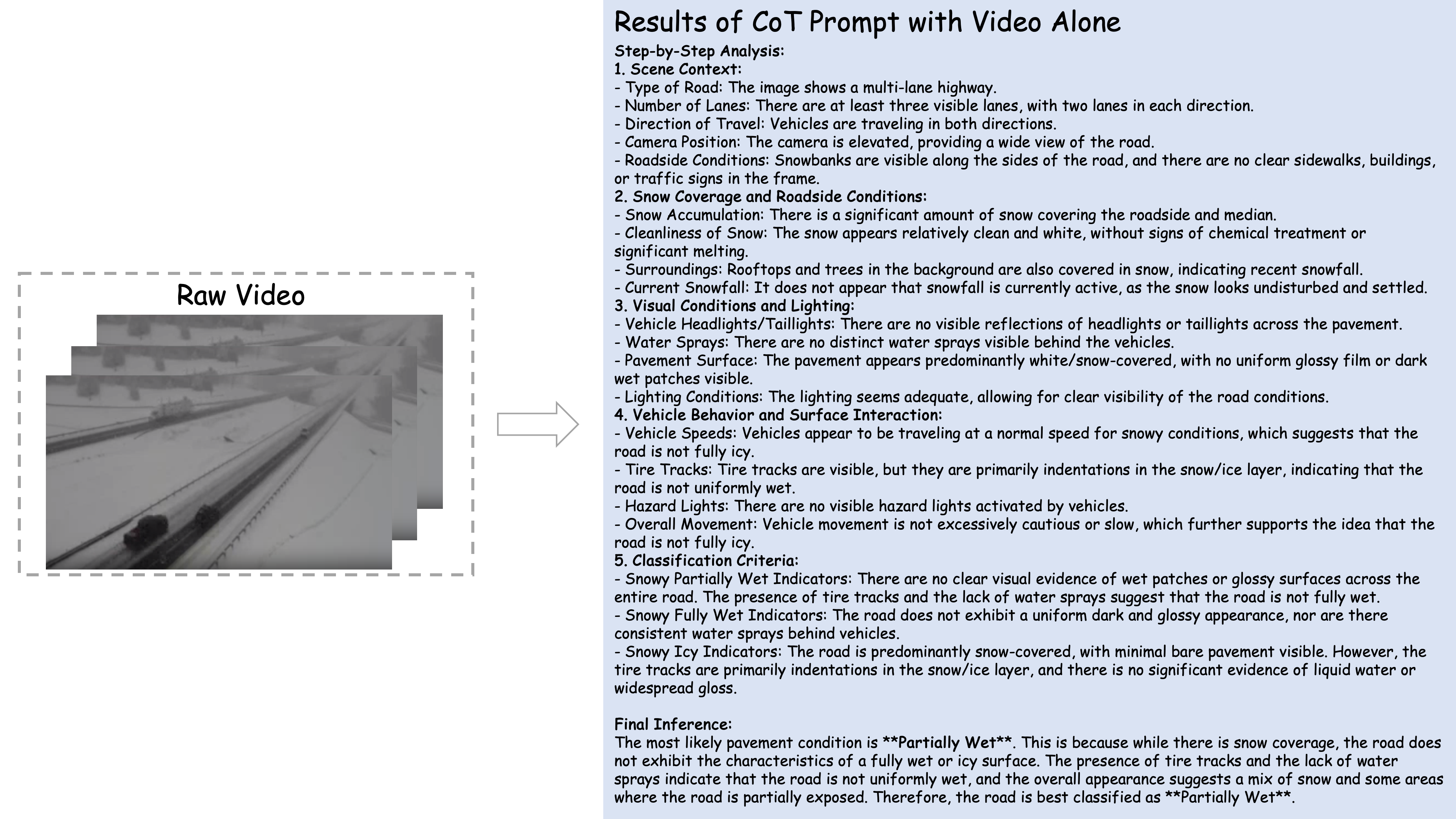}
    \caption{Pavement wetness level classification via CoT prompt without multimodal data under snowy fully wet condition.}
    \label{fig:wetness_snowy_fully_wet_wo_mm_data_result}
\end{figure}

\begin{figure}[!ht]
    \centering
    \includegraphics[width=0.95\linewidth]{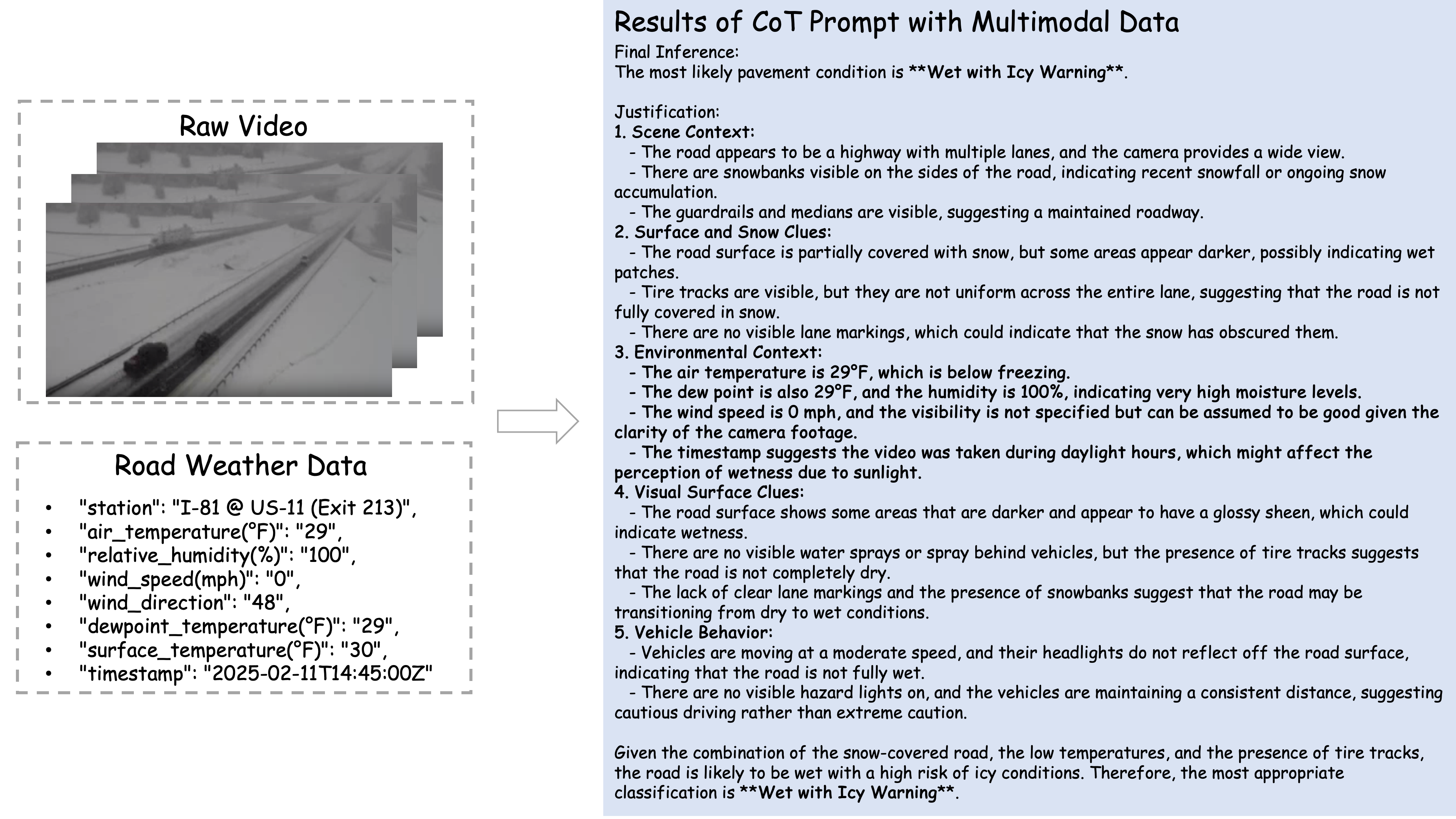}
    \caption{Pavement wetness level classification via CoT prompt with multimodal data under snowy fully wet condition.}
    \label{fig:wetness_snowy_fully_wet_w_mm_data_result}
\end{figure}

We utilize the multimodal pavement condition dataset to highlight the benefits of multimodal reasoning. As shown in Figures \ref{fig:wetness_snowy_fully_wet_wo_mm_data_result} and \ref{fig:wetness_snowy_fully_wet_w_mm_data_result}, while the model can perform basic scene analysis based on snow coverage and other visual cues, the inclusion of additional modalities substantially improves its ability to generate more comprehensive and accurate assessments. By incorporating data such as temperature, dew point, and humidity, the model was able to successfully identify icy conditions, a capability particularly valuable for downstream transportation safety applications.

\begin{figure}[!ht]
    \centering
    \includegraphics[width=0.95\linewidth]{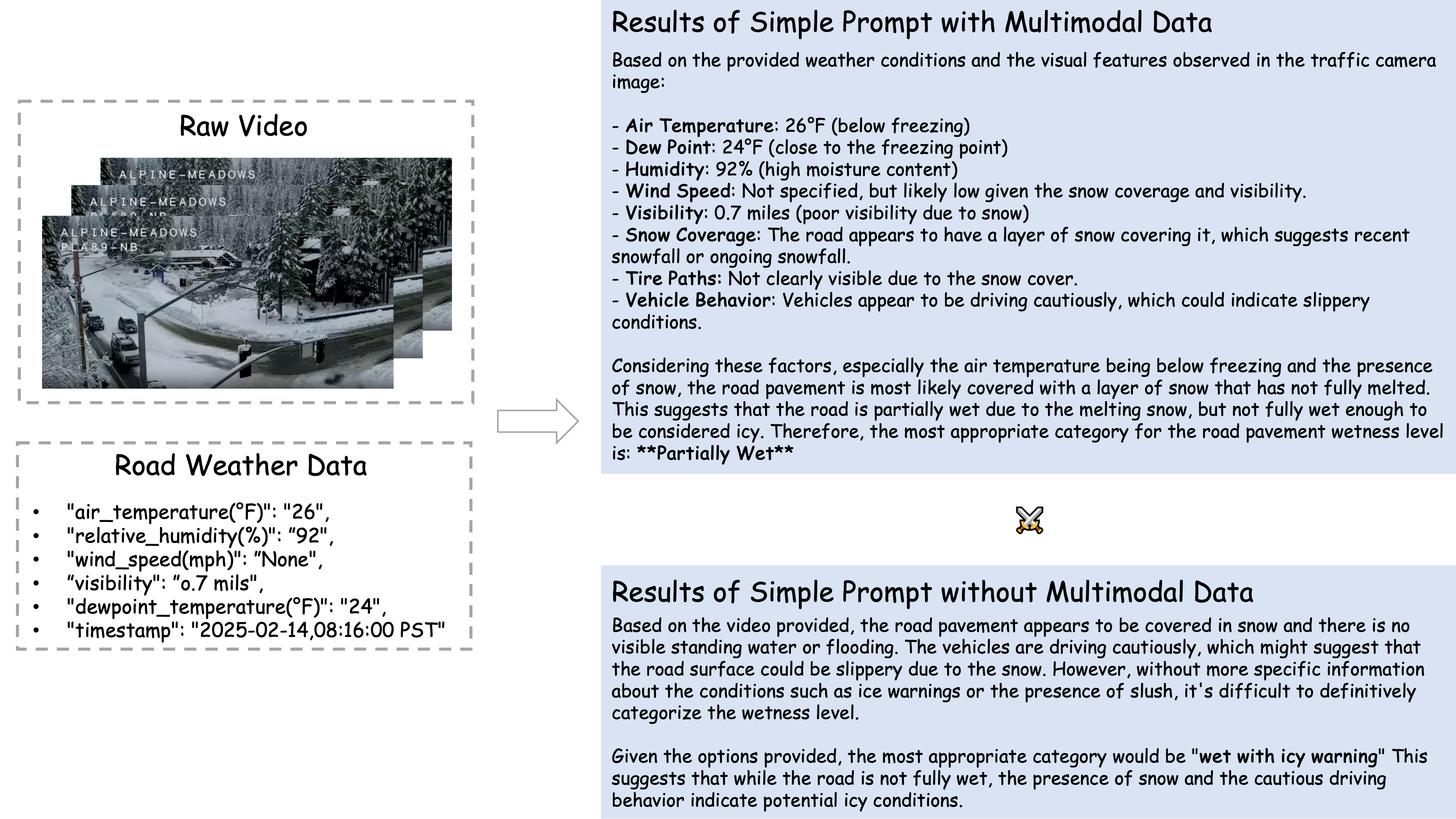}
    \caption{Comparison of pavement wetness level classification via simple prompt with/without multi-modal data under snowy wet with icy warning condition.}
    \label{fig:wetness_snowy_icy_simple_mm_comparison}
\end{figure}

We also evaluated the performance of multimodal reasoning on additional icy scenarios using the simple prompt, Surprisingly, even when the model could accurately infer conditions using only video data, the simple prompt sometimes produced incorrect results (see Figure \ref{fig:wetness_snowy_icy_simple_mm_comparison}). This suggests that while simple prompts may suffice in cases with strong and unambiguous visual cues, they fail when handling multimodal inputs, particularly when visual information is incomplete or ambiguous. Our results indicate that CoT (Chain-of-Thought) prompting is essential for robust performance with multimodal data (as shown in Table \ref{tab:ablation_combined}); otherwise, the model may struggle when processing environmental parameters without sufficient visual context. For instance, the model misclassified "snowy fully wet" scenes as "partially wet," and similar mistakes were observed for rainy conditions. These errors may stem from the model’s over-reliance on explicit visual cues (e.g., reflections, water spray, and road gloss) that are often subtle, inconsistent, or missing due to environmental and data limitations. In snowy scenes, slush and snow accumulation can obscure pavement texture, leading to misclassification of fully wet surfaces. Likewise, icy conditions can visually resemble fully wet roads, increasing the likelihood of false positives. In rainy scenarios, diminished sprays or weak headlight reflections may also cause false negatives. These findings highlight the challenges of relying solely on visual cues and underscore the importance of complementary modalities and thought-provoking prompt design for robust multimodal reasoning.

\vspace{1em}
These challenges are further exacerbated by two key factors: (1) the short duration of the video clips, ranging from only 4 to 7 seconds, which limits the availability of temporal cues such as sustained water trails or subtle vehicle motion dynamics, and (2) the varying video resolution, with many clips being of low-quality, hinders the model’s ability to detect fine-grained visual features essential for accurately reasoning about surface conditions.

\begin{table}[!ht]
    \caption{Accuracy comparison between simple and CoT prompts across input modalities (Video alone versus Multimodal data) using Agent 2 (QWEN 2.5-VL-7B model)}
    \label{tab:ablation_combined}
    \begin{center}
    \resizebox{\textwidth}{!}{%
        \begin{tabular}{l l c c}
            \hline
            \textbf{Input Type} & \textbf{Condition} & \textbf{Simple Prompt (\%)} & \textbf{CoT Prompt (\%)} \\
            \hline
            {Video alone} 
            & Rainy partially wet         & 37.25 & 58.82 (\textbf{↑21.57}) \\
            & Rainy fully wet             & 23.29 & 57.33 (\textbf{↑34.04}) \\
            & Rainy flooded               & 0.00  & 57.89 (\textbf{↑57.89}) \\
            & Snowy partially wet         & 20.00 & 100.00 (\textbf{↑80.00}) \\
            & Snowy fully wet             & 55.56 & 0.00 (\textbf{↓55.56}) \\
            & Snowy wet with icy warning  & 76.19 & 14.29 (\textbf{↓61.90}) \\
            & Sunny dry                   & 83.33 & 95.45 (\textbf{↑12.12}) \\
            \hline
            {Multimodal (Video and Sensor Data)} 
            & Snowy fully wet             & 10.00 & 100.00 (\textbf{↑90.00}) \\
            & Snowy wet with icy warning  & 14.29 & 100.00 (\textbf{↑85.71}) \\
            \hline
        \end{tabular}%
    }
    \end{center}
\end{table}

\section{Conclusions and Future Work}

In this work, we proposed a general multi-agent framework for comprehensive highway scene understanding. The framework leverages a large VLM (e.g., GPT-4o) to generate CoT prompts enriched with domain knowledge, which are then used to guide a smaller, efficient VLM (e.g., Qwen2.5-VL-7B) in reasoning over video inputs, with complementary modalities as applicable. This design enables robust performance across multiple core perception tasks, including weather classification, pavement wetness assessment, and traffic congestion detection.

To evaluate the effectiveness of the proposed framework, we curated three datasets. For the pavement wetness assessment task in particular, we constructed a multimodal dataset to demonstrate the benefits of multimodal reasoning. By leveraging carefully designed CoT prompts, the framework achieves significantly improved reasoning performance and substantial gains in overall accuracy. This zero-shot, multi-agent approach leverages domain knowledge through a large VLM and unlocks the potential of small VLMs, offering a scalable, cost-effective solution for diverse transportation applications. Our framework can be readily integrated with the abundant network of existing traffic cameras, enabling large-scale deployment. In rural regions, where traditional sensor coverage is sparse, our method supports strategic monitoring by focusing on high-risk locations such as sharp curves, flood-prone lowlands, or icy bridges. By continuously analyzing scene conditions at these targeted sites, the system enhances situational awareness and provides timely alerts even in disconnected environments. Additionally, the ability to automatically detect congestion and road hazards allows transportation agencies to efficiently screen regional or statewide traffic camera feeds and quickly identify problem areas without intensive manual review.

\vspace{1em}
Nonetheless, there remains substantial room for improvement. A promising direction is to distill or design a more compact VLM tailored to the target tasks. Such a lightweight model would be well-suited for edge deployment, facilitating the integration of advanced AI capabilities into existing traffic camera networks and enabling scalable, real-time intelligent scene understanding.

\section{Acknowledgements}
This research was supported by the U.S. Department of Transportation, Office of the Assistant Secretary for Research and Technology (OST-R), University Transportation Centers Program, through the Center for Regional and Rural Connected Communities (CR2C2) under Grant No. 69A3552348304. 

\section{Author Contributions}
Conception and design: J.J.Y. and Y.Y.; data processing: Y.Y. and N.X.; analysis and interpretation of results: Y.Y., N.X., and J.J.Y.; draft manuscript preparation: Y.Y. and N.X.; review and editing: J.J.Y.; visualization, Y.Y.; supervision, J.J.Y.; project administration, J.J.Y.; funding acquisition, J.J.Y. All authors have read and agreed to the published version of the manuscript.


\bibliographystyle{unsrt} 
\bibliography{references}  
\end{document}